\documentclass[conference]{IEEEtran}

\usepackage[dvipsnames]{xcolor}
\usepackage[latin9]{inputenc}
\usepackage{amsmath}

\usepackage{graphicx}
\usepackage{algorithm}
\usepackage{algorithmicx}
\usepackage{todonotes}
\usepackage[noend]{algpseudocode}
\usepackage{comment}
\usepackage{multirow}
\usepackage{subfigure}
\usepackage{ifthen}
\newcommand{\ru}    {\rule{0mm}{3.25mm}}
\newcommand{\ped}[1]{\scriptscriptstyle\mathrm{#1}}

\ifCLASSINFOpdf
\else
\fi

\makeatletter
\def\endthebibliography{%
	\def\@noitemerr{\@latex@warning{Empty `thebibliography' environment}}%
	\endlist
}
\makeatother

\usepackage{stfloats}
\begin{document}
\title{Incremental learning for the detection and classification of GAN-generated images}

\author{
\IEEEauthorblockN{Francesco Marra$^{\star}$, Cristiano Saltori$^{\dagger}$,
Giulia Boato$^{\dagger}$ and Luisa Verdoliva$^{\star}$} 
\IEEEauthorblockA{ $^{\star}$ {\em University Federico II of Naples, Italy} \\
$^{\dagger}$ {\em University of Trento, Italy}
\\ 
}
}

\maketitle

\begin{figure}[b]
\vspace{-0.4cm}
\parbox{\hsize}{\em
WIFS`2019, December, 9-12, 2019, Delft, Netherlands.
978-1-7281-3217-4/19/\$31.00 \ \copyright 2019 IEEE.
}\end{figure}

\begin{abstract}

Current developments in computer vision and deep learning allow to automatically generate hyper-realistic images, hardly distinguishable from real ones. 
In particular, human face generation achieved a stunning level of realism, 
opening new opportunities for the creative industry but, at the same time, new scary scenarios where such content can be maliciously misused. 
Therefore, it is essential to develop innovative methodologies to automatically tell apart real from computer generated multimedia,
possibly able to follow the evolution and continuous improvement of data in terms of quality and realism.
In the last few years, several deep learning-based solutions have been proposed for this problem, mostly based on Convolutional Neural Networks (CNNs).
Although results are good in controlled conditions, 
it is not clear how such proposals can adapt to real-world scenarios, 
where learning needs to continuously evolve as new types of generated data appear.
In this work, we tackle this problem by proposing an approach based on incremental learning for the detection and classification of GAN-generated images.
Experiments on a dataset comprising images generated by several GAN-based architectures 
show that the proposed method is able to correctly perform discrimination when new GANs are presented to the network.

\end{abstract}

\IEEEpeerreviewmaketitle
\section{Introduction}

Our perception of what is real and what is fake has evolved over time, hand-in-hand with advancing technology.
Nowadays, thanks to deep neural networks and more specifically to Generative Adversarial Networks (GANs) \cite{goodfellow2014generative} 
even a non-expert user can generate images or manipulate video sequences in a compelling way.
With a well-trained GAN, one can easily create a new photo from scratch \cite{Berthelot2017} and real-time rendering of photo-realistic facial animation is just around the corner.
Other methodologies allow to translate a given image into a new context or modify a specific attribute, such as hair color or facial expression \cite{Zhu2017,Choi2018}.
The level of photo-realism achieved by such methods has become impressive, especially for faces \cite{brock2018large, Karras2018}, such to deceive even human inspection.
Hyper-realism will soon become widely accessible and go beyond the primary purpose of entertainment.
This scenario opens up incredible opportunities in many fields like content production, education, health \& assisted living. 
However, the wide availability of possibly uncontrolled, distributed and real-time online generated data 
raises serious concerns on the trustworthiness of content 
which could be misused in various ways, e.g., to convey wrong or malicious information, or to bias people and influence social groups.
In this context, there is a fundamental need to develop techniques able to automatically discriminate generated from real content, 
and also able to follow the continuous evolution of image generation in terms of quality and realism, 
thus supporting human beings to preserve awareness of what is real and what is not.

\begin{figure}[t!]
    \centering
    \includegraphics[width=\columnwidth]{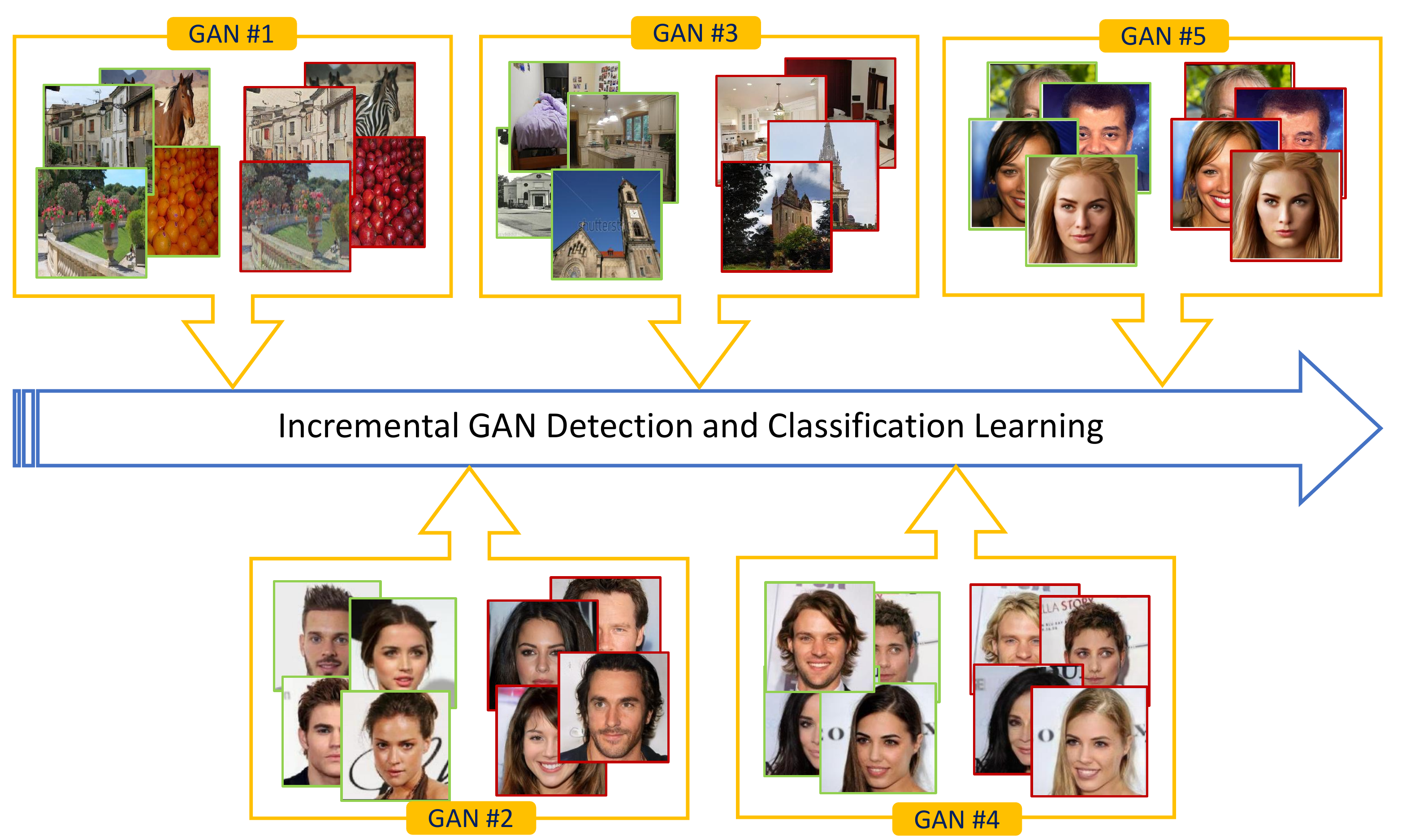}
    \caption{The GAN-incremental scenario considered in the paper. Our aim is to propose a strategy that is able to detect and classify new GAN generated images, without reducing performance on the previous ones. All images with green border are real while those with red borders are generated.
    }
    \vspace{-4mm}
    \label{fig:teaser}
\end{figure}

Several methods have been already proposed in the literature to detect whether an image is GAN-generated or not.
Some of them exploit specific facial artifacts,
like asymmetries in the colour of the eyes, or artifacts arising from an imprecise estimation of the underlying geometry,
especially on areas around the nose, the border of the face, and the eyebrows \cite{Matern2019}.
Color information is instead exploited in \cite{McCloskey2018, Li2018}.
In particular, \cite{McCloskey2018} proposes to use features shared by different GAN architectures,
based on the way they transform a multi-channel feature map into a 3-channel color image.
In \cite{Yang2019} the authors observe that 
the face configuration in synthetic images has different characteristics than in real ones, given the lack of global constraints. 
Hence, it is possible to train a SVM by simply using the locations of facial landmark points.

The presence of specific GAN artifacts suggests that each generated image may be characterized by a specific fingerprint
just like natural images are identified by a device-based signature (i.e. PRNU).
These artificial fingerprints have been studied in \cite{Yu2018, Marra2019},
they have been shown to characterize not only a given architecture but also a specific instantiation of it, 
e.g., a particular training on a given dataset.
Following this same path, in \cite{Albright2019} the problem of attributing a synthetic image to a specific generator is investigated in a white box setting, 
by inverting the generation process.

Approaches based on convolutional neural networks have proven to be very effective.
Several architectures have been proposed so far 
\cite{Marra2018, Nataraj2019, Xuan2019} 
showing a very good accuracy in detecting GAN-generated images, even after compression.
The main problem is that new GAN architectures for generating synthetic data are proposed by the day,
requiring the detector to be either re-trained on larger and larger training sets, or fine-tuned on them.
The first solution requires to store very large datasets of images, which is not feasible as data increase continuously, 
while the latter does not provide satisfactory results unless specific solutions are devised \cite{Cozzolino2018FT}.
In fact, a simple fine-tuning on a new dataset tends to destroy the information acquired in the previous training, 
a well-know phenomenon called \textit{catastrophic forgetting} \cite{kirkpatrick2017overcoming}.

In this work we face this challenging problem 
and propose a multi-task incremental learning method specifically targeted to GAN-generated image detection.
Specifically, we want to propose an approach able to 
to detect and classify new GAN generated images, without worsening the performance on the previous ones (see Fig.\ref{fig:teaser}).
The method is inspired by the Incremental Classifier and Representation Learning (iCaRL) technique, recently proposed in \cite{rebuffi2017icarl} for object classification, which adapts to new classes without forgetting the old ones thanks to a small memory of suitable images.
To exploit such an approach in our context, we consider a multi-task problem involving both GAN-image classification and detection,
and add a new binary loss term to the existing classification loss. 
In this way, we perform both the detection and classification of the GAN architecture simultaneously.
In the next sections, we describe the iCaRL method for incremental learning,
then present our method, experimental results, and finally draw conclusions.

\section{Incremental Classifier using iCARL}\
\label{sec:icarl}

\renewcommand{\P}{{\cal P}}
\newcommand{\X}{{\cal X}}
There is a growing literature on incremental classifiers, that is,
classifiers which adapt to an increasing number of classes without re-training on the whole dataset \cite{rebuffi2017icarl,Javed2019icarlrevised,Lopez-Paz2017}.
Unlike in the conventional scenario, where all data from all classes of interest are available at training time, 
in the class-incremental scenario, new classes keep appearing over time, with the associated training data.
Therefore, an {\it ad hoc} class-incremental learning strategy is necessary, which allows updating the classifier as new sets of classes keep appearing.
Crucial for the success of such strategies is a mechanism to avoid {\em forgetting} previous classes while adapting to the new ones.
We selected a successful algorithm for incremental learning, iCaRL \cite{rebuffi2017icarl}, based on a general purpose CNN-based classifier and a set of {\em exemplar images}, $\P$, extracted evenly from all available classes.
Notably, the CNN is trained and updated only to work as a feature extractor,
while the actual classification is performed based on the feature vectors associated with the exemplar images.
The exemplar images themselves are the key to avoid forgetting previous classes.
However, to limit memory resources, their number, also called memory budget $M=|{\cal P}|$, is not allowed to grow over time, and hence this is the main parameter of the algorithm.
Let us briefly describe iCaRL by separating the initialization and the updating phases.

{\em Initialization:}
let $s$ classes be given, with associated training sets $\X_1,\ldots,\X_s$.
A CNN is trained on such data to minimize the classification error.
When the training is over, the output of a suitable layer is used to associate a unit-norm feature vector, $\phi(x)$, with each input image $x$.
For each class, $y=1,\ldots,s$, $M/s$ images are selected to form the class-$y$ exemplar set, $\P_y$, and a class template vector is computed as the average of the corresponding feature vectors
\begin{equation}
    \mu_y = \frac{1}{|P_y|} \sum_{x_i \in P_y}\phi(x_i)
\end{equation}
These template vectors are eventually used to classify test images, according to a minimum distance rule:
\begin{equation}
    \hat{y} = \arg\min_{y=1,\ldots,s} ||\phi(x)-\mu_y||
\end{equation}

{\em Updating:}
let $t-s$ new classes be given, with associated training sets $\X_{s+1},\ldots,\X_t$.
iCaRL performs a three-step updating procedure:
\begin{enumerate}
\item   the weights of the CNN are updated, using only the new training sets and the set of exemplar images $\P$;
\item   exemplar sets are created for each of the new classes;
\item   a suitable number of images are discarded from old exemplar sets to keep a fixed memory budget $M$.
\end{enumerate}
These steps are detailed in the following.

1) CNN updating.
The weights $\Phi$ of the CNN are fine-tuned on the union of all new training sets $\X=\bigcup_{y=s+1}^t \X_y$ 
and on the union of all existing exemplar sets $\P=\bigcup_{y=1}^s \P_y$.
Since $M=|\P|$ is much smaller than the aggregated size of the old training sets, this process is much less memory and computation intensive than full re-training.
In particular, the updating relies on a suitable loss function

\begin{equation}
    \ell_{\rm iCaRL}(\Phi) = (1-\gamma) \ell_{\rm class}(\X,\Phi) +\gamma \ell_{\rm distill}(\P,\Phi)
\label{eq:icarlLoss}
\end{equation}
which balances, through a suitable weight $\gamma$, 
a {\em classification} term computed on the new training set, and a {\em distillation} term \cite{hinton2015distilling} concerning the exemplar set.
The classification term is the usual cross-entropy loss calculated on the new data 
\begin{equation}
    \ell_{\rm class}(\X,\Phi) = \sum_{x_i \in X} \delta_{y=y_i} \log g_y(x_i) + \delta_{y\neq y_i} log(1-g_y(x_i))
\end{equation}
where $g_y(\cdot)$ is the classification score for class $y$. The distillation term, is the \textit{KL-divergence} loss with temperature $T$, as proposed in \cite{hinton2015distilling}:     
\begin{equation}
    \ell_{\rm distill}(\P,\Phi) = \sum_{x_i \in \P} T^2 D_{KL}(g^T(x_i)||\tilde{g}^T(x_i) ) 
\end{equation}
where $\tilde{g}(\cdot)$ is the old classifier, that is, before the current updating phase. 
By including this term we are preserving outputs for the previous classes in $\P$ while we are learning parameters that are discriminative for the new classes.

2) Creation of new exemplar sets.
The exemplar set $\P_y$ is the major actor in preserving a synthetic representation of class $y$, thus preventing catastrophic forgetting.
Therefore, the selected exemplar vectors should be themselves good representative of the class,
also because the classification of test images will depend on their average.
For this reason, after computing the grand average over all feature vectors of the class
\begin{equation}
    \eta_y = \frac{1}{|\X_y|} \sum_{x \in \X_y} \phi(x)
\end{equation}
the selected exemplar vectors are simply those characterized by the smallest distance with respect to this average.
To respect the constraint on the memory budget, and have an equal number of vectors for each class, $M/t$ vectors are selected.

3) Reduction of old exemplar sets.
Following the very same reasoning as before, for existing exemplar sets, $M/s-M/t$ vectors must be discarded, and they are chosen as the farthest from the class template vector.

\section{Proposed method}
\label{sec:gan_icarl}

\begin{figure*}[!tbp]
  \centering
    \includegraphics[trim=0.0cm 9cm 0cm 10cm,width=\textwidth]{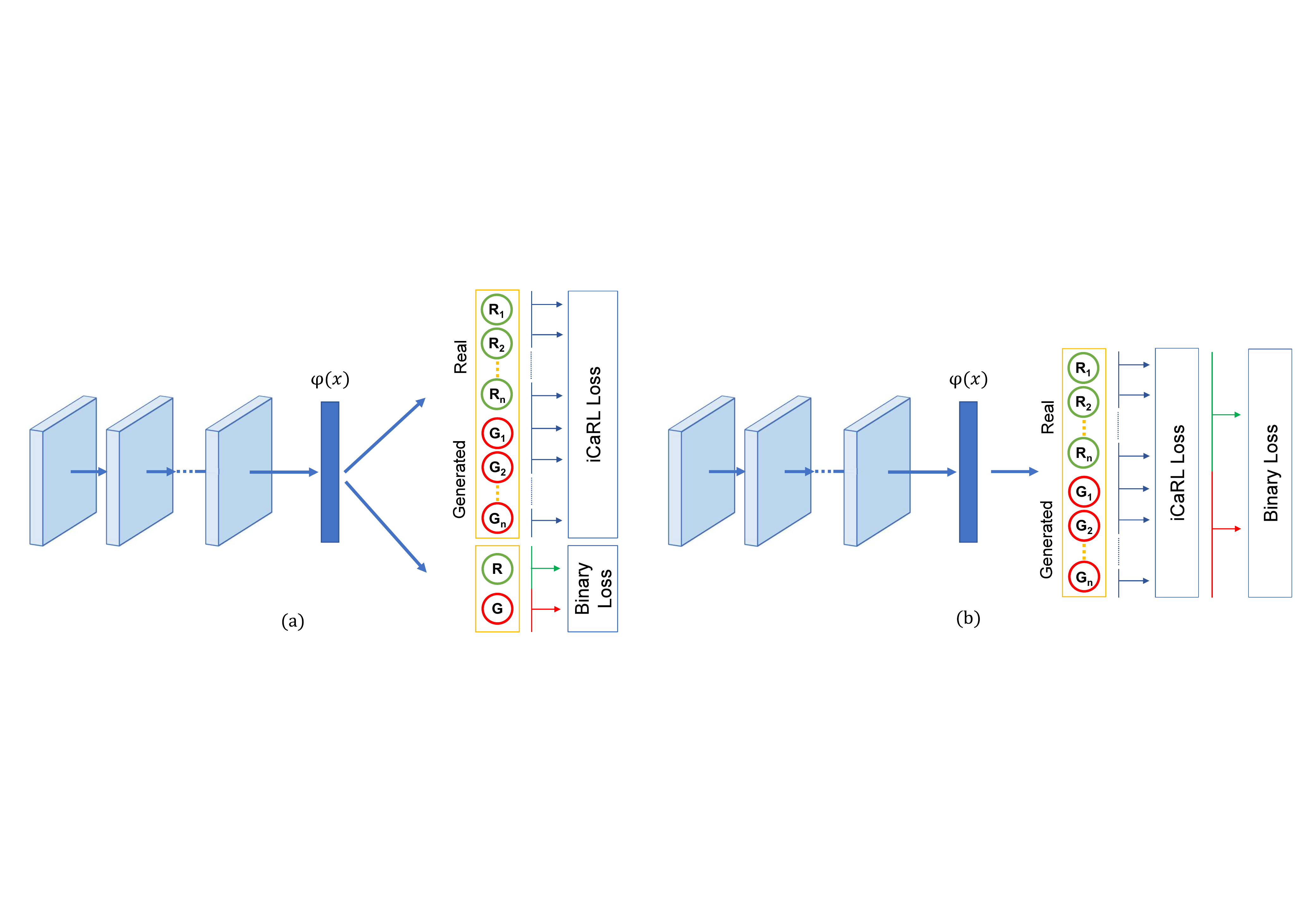}
    \caption{The two proposed methods: (a) MT-MC adds a new separate binary classifier trained jointly with the original one, and (b) MT-SC includes an additional binary loss to the iCaRL classifier.}
  \vspace{-4mm}
  \label{fig:GAN-iCaRL}
\end{figure*}

From the previous section, 
it is clear that iCaRL represents a precious tool to build a classifier 
that is seamlessly upgraded to deal with new classes as they appear, without losing the capability to correctly classify old ones.
In this work, we depart from the standard object classification task, for which iCaRL was originally conceived,
and adapt it to the detection and classification of GAN-generated images.
To this end, we propose two multi-task versions of the original iCaRL algorithm, with suitable training procedures.

Now, the detection of GAN-generated images is a {\em binary} problem, with only two classes: real and GAN-generated.
Therefore, in principle, it has little to do with the incremental class scenario.
In this paper, however, we will address the detection problem by leveraging the classification ability of the system.
That is, based on the knowledge of existing GAN-architectures, we will train a suitable classifier for them,
and eventually label an image as GAN-generated if the classifier decides for any of the possible GAN classes.
Under this perspective, incremental learning becomes the perfect tool to deal not only with existing, but also with upcoming GAN architectures\footnote{By so doing, we are sort of bypassing the detection problem.
However, to the best of our knowledge, no ``GAN smoking gun'' has been identified to date which enables reliable direct detection,
hence we believe this approach is not only legitimate but fully sensible and conservative.}.

Following this approach,
we treat images from different GAN architectures as coming from different classes, addressing the classification and detection problems jointly. 
Note that to each architecture we actually associate {\em two} classes, 
GAN-generated ($G$) and real ($R$), with their corresponding training sets.
So, for the $i$-th architecture we have two distinct sets ${\cal X}^G_i$ and ${\cal X}^R_i$,
and all sets of real images are disjoint and possibly unrelated.
When a new class of GAN-generated images appears, that is, a new architecture, 
we use incremental learning to update both classifier and detector.
Therefore, we are addressing a forensic multi-task (detection and classification) problem.
In particular, we propose two solutions based on the iCaRL algorithm.
\begin{enumerate}
\item   Our first proposal, represented graphically in Fig.\ref{fig:GAN-iCaRL}.a, 
        consists in using a distinct binary classifier $d(\cdot)$, that is, a detector, in parallel with the incremental classifier $g(\cdot)$.
        The detector decides on whether the input is GAN-generated or not and is characterized by a suitable binary loss.
        Hence, we will call this solution, Multi-Task Multi-Classifier (MT-MC).
\item   Following a different line of thinking, 
        we can regard the detection task as the juxtaposition of multiple binary classification tasks. 
        Therefore, in our second solution, represented graphically in Fig.\ref{fig:GAN-iCaRL}.b,  
        the structure is left unchanged, and detection is a direct by-product of classification.
        However, we include again a binary loss term in the classifier to push the CNN weights towards solving the detection problem.
        We refer to this second proposal as Multi-Task Single Classifier (MT-SC).
\end{enumerate}

In the following, without loss of generality, we describe the two variants 
with reference to the case in which a single GAN is added at the current step, say $n$.
Moreover, we use ${\cal X}_n=({\cal X}^G_n,{\cal X}^R_n)$ to indicate the two datasets associated with the new GAN analyzed at step $n$, 
while ${\cal P}$ indicates, as usual, the union of all exemplar sets, both GAN and real. 

In the Multi-Task Multi-Classifier (MT-MC) variant, 
a separate binary detector is used, characterized by its own loss
\begin{equation}
\small{
 \ell_{\ped{BL}}(\X_n,\Phi) = \sum_{x_i \in {\cal X}_n} \delta_{\scriptscriptstyle\mathrm{Y=G}} d(x_i) + \delta_{Y=R} (1-d(x_i))
}
\end{equation}
where $d(x_i)$ is the detector output and $Y$ is the binary label of interest.
The network is trained by using together  this loss and the usual iCARL loss, 
in a classical multi-task learning fashion, with aggregated loss:
\begin{equation}
    \small{
    \begin{aligned}
    \ell_{\ped{MTMC}}(\Phi) & = \ell_{\ped{iCaRL}}(\P,\X,\Phi) + \lambda \ell_{\ped{BL}}(\X_n,\Phi)
    \end{aligned}
    }
    \label{eq:loss_function_MT_MC}
\end{equation}
so as to boost both the classification and detection performance while the number of classes grows.

In Multi-Task Single-Classifier (MT-SC), instead, both the detection and the classification tasks are managed by the same classifier $g(\cdot)$. 
To enforce the desired additional constraint, we update the loss by including  a binary cross-entropy term
\begin{equation}
\label{eq:loss_function_MT_SC}
    \small{
    \begin{aligned}
    \ell_{\ped{MTSC}}(\Phi) & = \ell_{\ped{iCaRL}}(\P,\X_n,\Phi) + \lambda \ell'_{\ped{BL}}(\P,\X_n,\Phi) \\
    \end{aligned}
    }
\end{equation}
The additional loss is computed by taking into account the activations, $g(\cdot)$, of all the classes, separately for the GAN and real classes.
Formally,
\begin{equation}
\small{
    \begin{aligned}
    \ell'_{\ped{BL}}(\P,\X_n,\Phi) = \sum_{x_i \in (\P,\X_n)} \delta_{\scriptscriptstyle\mathrm{Y=G}} d_G(x_i) +  \delta_{Y=R} d_R(x_i) \\
    \end{aligned}
    }
\end{equation}
where
\begin{equation}
\small{
    \begin{aligned}
    d_{\ped{G}}(x_i) = \hspace{-3mm} \sum_{y \in \{GAN\}}  \hspace{-3mm}\log g_y(x_i) \quad
    d_{\ped{R}}(x_i) = \hspace{-3mm} \sum_{y \in \{Real\}} \hspace{-3mm}\log g_y(x_i)
    \end{aligned}
    }
\end{equation}
Here, with $y \in \{GAN\}$ we mean all classes corresponding to the various GAN architectures,
while $y \in \{Real\}$ indicates the associated real classes.

Then, these two terms will back-propagate jointly during the training phase.

\section{Experimental Results}

To validate the proposed method we perform experiments on a
publicly available GAN-dataset\footnote{http://www.grip.unina.it/download/DoGANs/} described in Sec.\ref{sec:dataset}.
Then, we define the parameter setting and the network configuration (Sec.\ref{sec:setup}), and describe experimental results for both GAN detection (in Sec.\ref{sec:detectionresults}) and classification (Sec.\ref{sec:classificationresults}) in comparison with state-of-the-art approaches.

\subsection{GAN-image dataset}
\label{sec:dataset}

\begin{figure}[t!]
    \centering
    \includegraphics[width=\columnwidth]{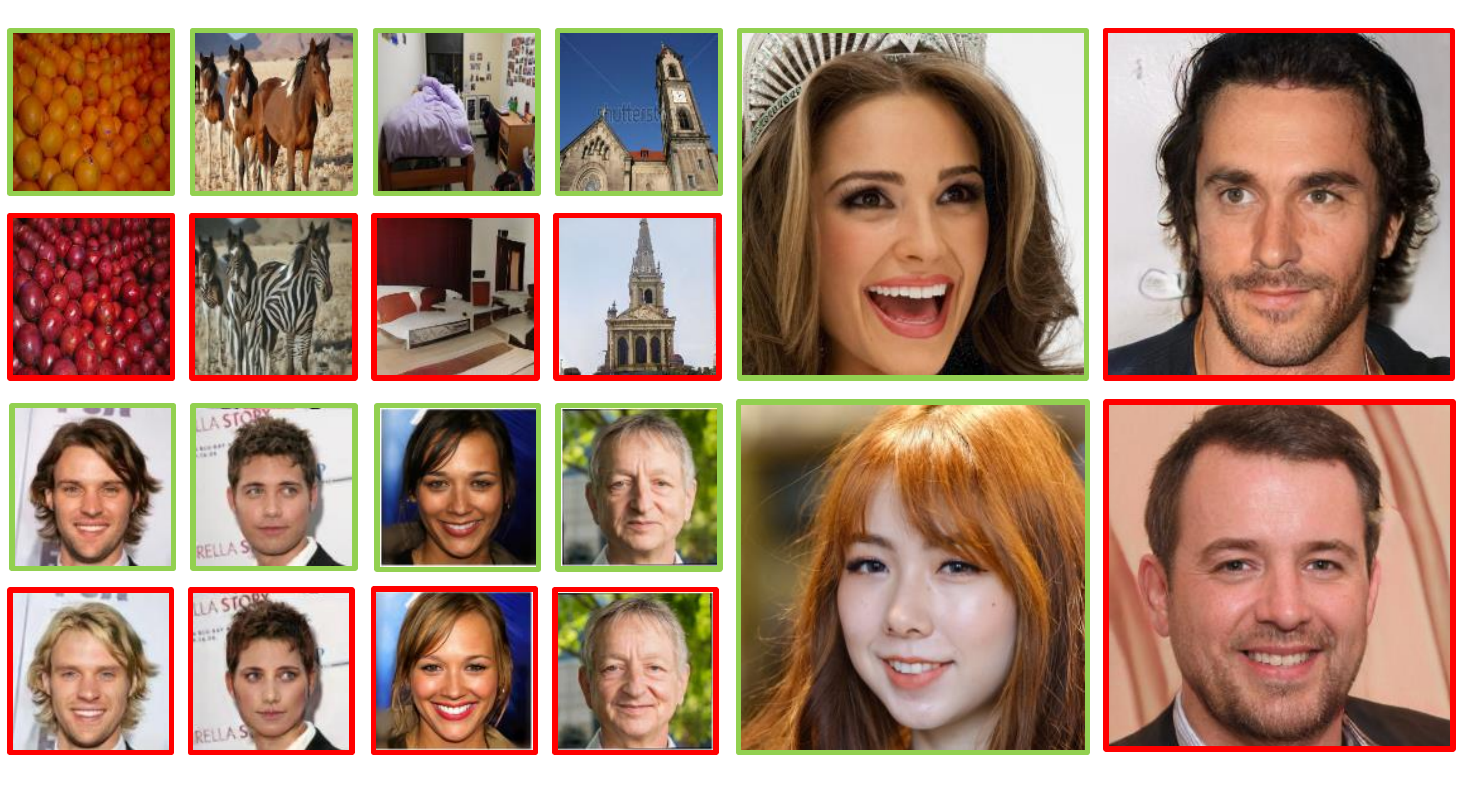}
    \caption{Representative images from our GAN dataset. All images with green border are real while those with red borders are generated.}
    \label{fig:images}
\end{figure}

The dataset is built using five different well-known state-of-the-art generative architectures, namely: 
CycleGAN \cite{Zhu2017}, ProGAN \cite{Karras2018}, both the $256\times256$ and the $1024\times1024$ versions, Glow \cite{kingma2018glow} and StarGAN \cite{Choi2018}.
Images generated from each architecture have been divided into a training, a validation and a test set with 3600, 2400 and 2400 per-class images, respectively.
In Tab. \ref{tab:dataset} we present all the information about the datasets, while in Fig. \ref{fig:images} we show some of the real (green border) and generated (red border) images.
Note that pristine samples selected for different architectures do not overlap, so as to avoid any form of polarization. 
We also make sure that styles used in training and validation are not present in the test sets.

\begin{table}[t]
  \centering
    \caption{Incremental-GAN dataset used in our experiments.}
    \begin{tabular}{c|c|c|c}
    \textbf{GAN} & \textbf{Method} & \textbf{Train} & \textbf{Test} \\
    \hline
    \multirow{7}[2]{*}{1} & \multirow{7}[2]{*}{Cycle-GAN} 
          & apple2orange & \multicolumn{1}{l}{photo2ukiyoe} \\
          &       & orange2apple & photo2vangogh \\
          &       & horse2zebra &  \\
          &       & zebra2horse &  \\
          &       & monet2photo &  \\
          &       & photo2cezanne &  \\
          &       & photo2monet &  \\
    \hline
    \multirow{3}[2]{*}{2} & \multirow{3}[2]{*}{ProGAN (256)} 
          & bedroom & kitchen \\
          &       & bridge & tower \\
          &       & churchoutdoor &  \\
   \hline
    \multirow{3}[2]{*}{3} & \multirow{3}[2]{*}{ProGAN (1024)} 
    &  &  \\
    && CelaebA-HQ & CelaebA-HQ \\
    &&  &  \\
    \hline
    \multirow{3}[2]{*}{4} & \multirow{3}[2]{*}{Glow} & Black Hair  & Male \\
          &       & Blond Hair & Smiling \\
          &       & Brown Hair &  \\
    \hline
    \multirow{3}[2]{*}{5} & \multirow{3}[2]{*}{starGAN} & Black Hair & Male \\
          &       & Blond Hair & Smiling \\
          &       & Brown Hair &  \\
    \end{tabular}%
  \label{tab:dataset}%
\end{table}%

\subsection{Experimental setup}
\label{sec:setup}

All the networks were trained using patches of $256\times256$ pixels randomly cropped from the training images, while in the test phase central cropped patches were used.
XceptionNet has been initialized using the off-the-shelf pre-trained weights on ImageNet and then trained with the ADAM gradient-based optimization scheme setting a learning rate of 0.001, a batch-size of 64 and default values for the moments.
At each class increment, we train the network until the loss, evaluated on the validation, does not improve for 5 consecutive epochs.
For all the methods under comparison we used the same ADAM optimizer with a learning rate estimated empirically from the validation set for each model. All the image pixel values are scaled to  $[-1,1]$ as pre-processing.

For the proposed methods we need to set two parameters present in the loss function, eq. \ref{eq:loss_function_MT_MC} and \ref{eq:loss_function_MT_SC}.
The first parameter is the regularization term $\lambda$ on the additional binary loss, while the second parameter is the distillation temperature $T$.
In order to select the best ones, we fix the memory budget $M=512$ and select the first three GAN architectures of Tab. \ref{tab:dataset}.
The results are presented in Tab. \ref{tab:ablation}
and suggest to use a temperature $T=2$ for both the methods, a regularization $\lambda=1$ for the MT-MC and a $\lambda=0.5$ in the MT-SC.
Moreover, in all  experiments $\gamma$ (eq. \ref{eq:icarlLoss}) is set to $0.5$.

\subsection{GAN-detection results}
\label{sec:detectionresults}

In this section we asses the detection performance of the proposed method in the GAN-incremental scenario, hence feeding the network using one GAN a time and keeping in memory only a limited number $M$ of samples from the previous GANs.

\begin{table}[t!]
  \centering
  \caption{GAN-image detection accuracy of the MT-SC and MT-MC iCaRL, after the third GAN added, with a memory budget of 512}
    \begin{tabular}{c|l|r|r|r}
          \ru &       & \multicolumn{1}{c}{$\lambda$=0.25} & \multicolumn{1}{|c|}{$\lambda$=0.5} & \multicolumn{1}{c}{$\lambda$=1} \\ \hline
    \ru \multirow{3}[0]{*}{MT-SC}   & T=1   & 96.18 & 95.75 & 95.75 \\
    \ru                             & T=2   & 96.36 & \textbf{97.06} & 96.15 \\
    \ru                             & T=3   & 96.89 & 96.24 & 95.79 \\ \hline
    \ru \multirow{3}[0]{*}{MT-MC}   & T=1   & 91.25 & 87.36 & 89.70 \\
    \ru                             & T=2   & 90.70 & 91.71 & \textbf{92.54} \\
    \ru                             & T=3   & 91.78 & 91.13 & 92.31 \\
    \end{tabular}%
  \label{tab:ablation}%
\vspace{-4mm}

\end{table}%

For comparison we selected some state-of-the-art GAN detectors proposed recently. 
The first one is XceptionNet, the same architecture used in our iCaRL experiments, that shows good results in \cite{Marra2018} for GAN detection. 
Then, we compared the two training procedures proposed in \cite{Xuan2019}, 
where the discriminator of DCGAN \cite{radford2015unsupervised} is used as classifier 
and two different pre-processing procedures are used during the training phase, that is, Gaussian noise and Gaussian blur. We will refer to them as $M_{Gn}$ and $M_{Gb}$, respectively. 
Finally, we also considered the network proposed in \cite{Nataraj2019} that uses the co-occurrence matrix as input.

Since none of these networks has been proposed in a class-incremental scenario, 
to be as fair as possible we fine-tune them using all the data from the new classes and the limited exemplar set $\mathcal{P}$ available.
We also compare our proposal with the basic iCaRL version (that is, Single Task and Single Classifier) 
and two more strategies proposed in \cite{Javed2019icarlrevised}, referred to as S-Classifier and GS-Classifier.
Again, to ensure a fair comparison, we use XceptionNet as feature extractor for all these methods.

First of all, we evaluate the detection performance as a function of the memory budget $M$, considering also the extreme cases where $M=\infty$ (all the training images of the previous GAN-architectures are kept) and $M=0$ (the network does not keep any image from the past). 
In Tab.\ref{tab:BinaryAccuracy} we report the accuracy evaluated using the whole test set (all 5 architectures). 
For large memory budget, we observe a very good performance for most methods even in the absence of any incremental-learning strategy. 
When the memory budget decreases, instead, our proposal exhibits a large performance gain over standard methods and a smaller but consistent gain also over basic incremental-learning algorithms.

\begin{figure}[t]
    \centering
    \includegraphics[width=\columnwidth]{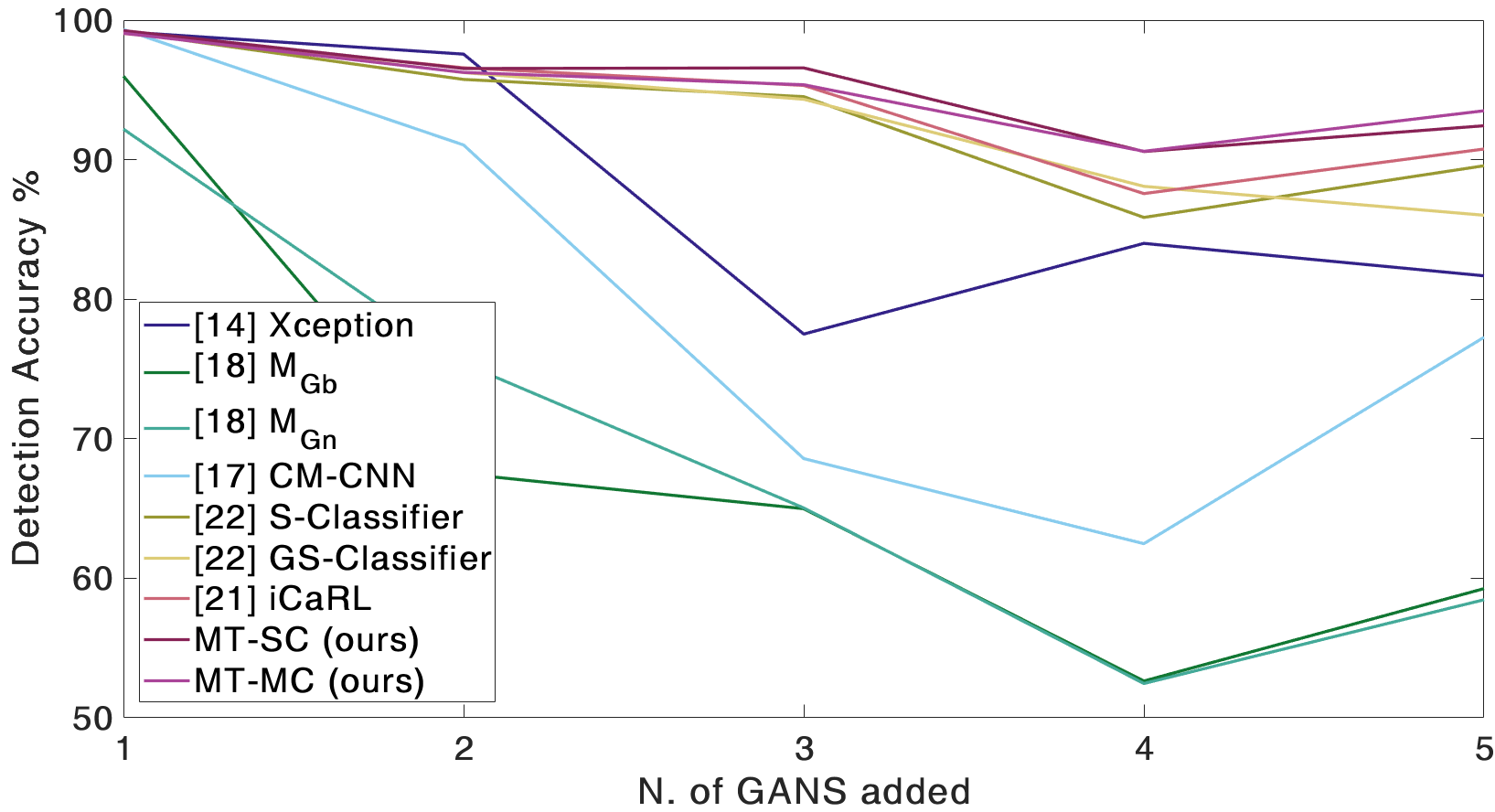}
    \caption{Detection accuracy on previous and current tasks increasing the number of GANs considered.
    All the methods have memory budget $M=256$.
    }
    \label{fig:permemor}
  \vspace{-4mm}

\end{figure}

In Fig.\ref{fig:permemor}, we evaluate the performance when a new GAN appears for a fixed memory budget $M=256$. 
The detection accuracy is computed on the old classes and the current one. 
It appears that when there is only one GAN to detect, all methods achieve a very good performance.
When the number of architectures increases, however, most techniques show large performance impairments.
Our proposals, instead, keep ensuring a detection accuracy over 90\%. 
It is worth noting that some curves increase sharply when the fifth GAN (StarGAN) is added. 
This is probably due to the fact that the corresponding pristine class includes images taken from the same dataset (CelebA-HQ) used for other architectures.
Although images are all distinct, the network improves its ability to recognize real faces.
\begin{table}[t]
  \centering
  \caption{Gan-image detection accuracy after the training on the last GAN using different memory budgets.}
    \begin{tabular}{l||c|c|c|c|c|c}
            & \multicolumn{6}{c}{Memory Budgets} \\ \hline
                                                & $\infty$  & 1024  & 512   & 256   & 128 & 0 \\ \hline
 \ru    \cite{Marra2018} Xception               & 99.10  & 93.97 & 87.39 & 81.67 & 79.95 & 65.55 \\
 \ru    \cite{Xuan2019} $M_{Gb}$                & 69.46  & 60.58 & 59.95 & 59.25 & 56.33 & 55.27 \\
 \ru     \cite{Xuan2019} $M_{Gn}$               & 67.82  & 59.03 & 57.26 & 58.46 & 55.15 & 54.14 \\
 \ru     \cite{Nataraj2019} CM-CNN              & 91.80 & 80.85 & 78.15 & 77.27 & 64.83 & 49.75 \\ \hline
 \ru     \cite{Javed2019icarlrevised} S-Classifier & -  & 96.05 & 92.52 & 89.56 & 84.43 & 64.38 \\
 \ru   \cite{Javed2019icarlrevised} GS-Classifier  & -  & 95.74 & 91.88 & 86.02 & 84.16 & 50.89 \\
 \ru    \cite{rebuffi2017icarl} iCaRL              & 97.43 & 95.51 & 93.63 & 90.76 & 89.81 & 70.59 \\ \hline
 \ru    MT-SC (ours)                               & 97.76 & 97.22 & 96.37 & 92.42 & 92.80 & 69.15 \\
 \ru    MT-MC (ours)                            & 99.37 & 94.50 & 95.36 & 93.50 & 86.47 & 67.71 \\
    \end{tabular}
  \label{tab:BinaryAccuracy}
\end{table}

\begin{table}[ht!]
  \centering
  \caption{Accuracy for the proposed methods after last training on Style-GAN using different memory budgets.}
    \begin{tabular}{l||c|c|c|c|c|c}
\ru          & \multicolumn{5}{c}{Memory Budgets}                  \\ \hline
\ru          & $\infty$  & 1024  & 512   & 256   & 128  & 0  \\ \hline
\ru    MT-SC (ours) &  98.28 & 97.39 & 95.64 & 94.32 & 92.77 &  63.67 \\
\ru    MT-MC (ours) &  97.85 & 96.34 & 95.37 & 93.03 & 89.89 &  70.25 \\
    \end{tabular}%
  \label{tab:adding}%
\end{table}%
Finally, we carried out a further experiment by adding a new GAN-architecture recently proposed in \cite{karras2019style}, called Style-GAN. 
In Tab.\ref{tab:adding} the detection results confirm that it is possible to increment the capability of the detector without affecting its performance.

\subsection{GAN-classification results}
\label{sec:classificationresults}

\begin{figure}[ht]
    \centering
    \includegraphics[width=0.75\columnwidth]{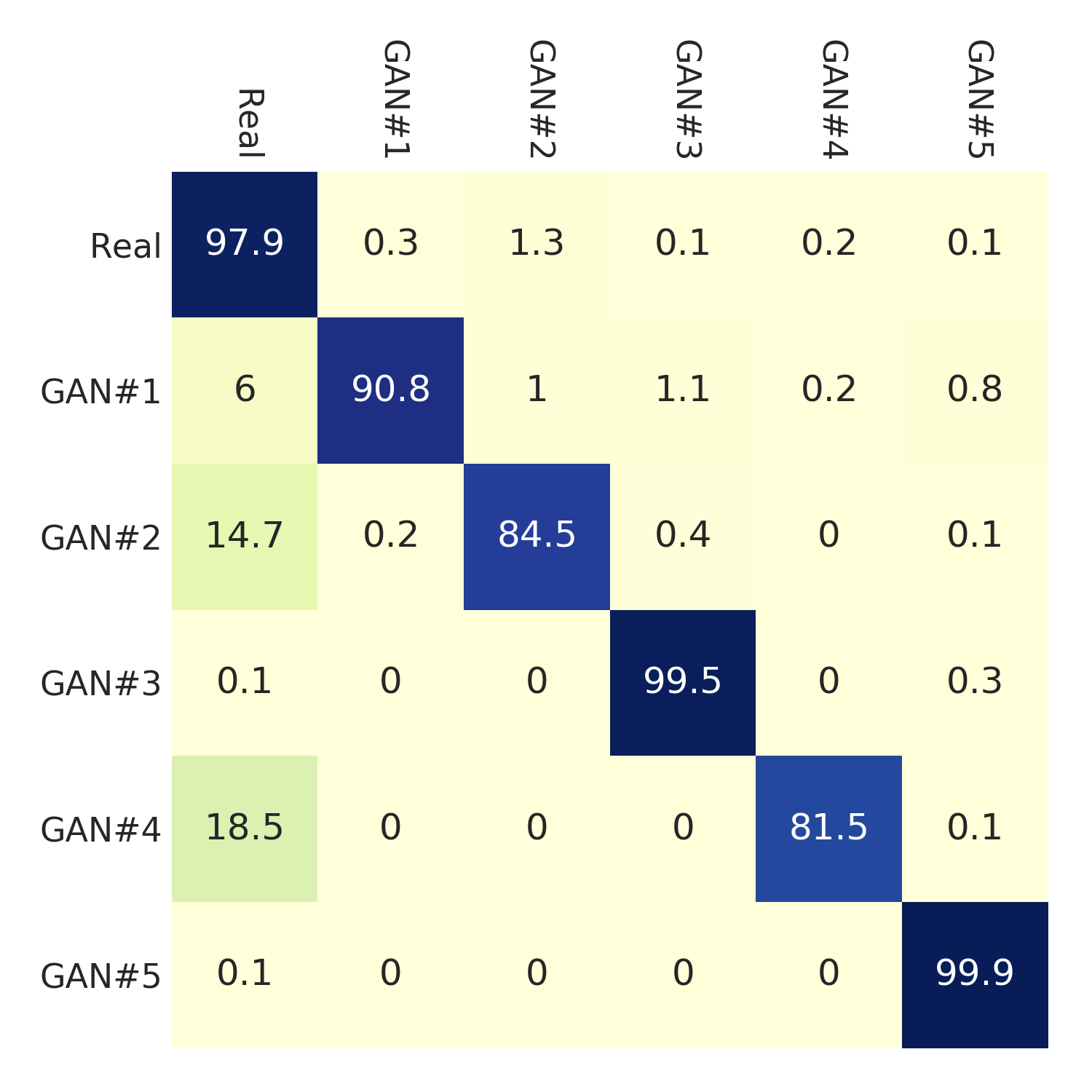}
    \caption{Confusion Matrix for classification of MT-MC with $M=256$ after the training on the last GAN.}
    \label{fig:cmatrixx}
    \vspace{-4mm}
\end{figure}

Since our proposal can also classify the specific type of GAN architecture, we analyze results also for this specific task.
In Fig.\ref{fig:cmatrixx} we show the confusion matrix produced by MT-SC with a memory budget $M=256$.
The classification performance is very good (accuracy above 90\%) for almost all the architectures and a bit worse for ProGAN and Glow (accuracy above 80\%). 
This kind of analysis is very important under a forensic point of view in order to identify the specific type of architecture used for image generation.

\section{Conclusion}

In this paper we address the problem of telling apart real images from images generated through adversarial learning. 
State-of-the-art CNNs exhibit a good performance when tested on the same type of images they were trained on, but often fail in detecting images generated by a different architecture.
To overcome this problem, we propose an incremental-learning strategy, building upon a method recently proposed for object classification.
Experiments show that our proposal is able to detect images generated by new GANs without reducing the performance on previous ones. 
In future work we will face the more challenging problem of detecting images generated by new GANs 
with no information on the architecture that generated them.

\section*{Acknowledgment}

We gratefully acknowledge the support of this research by a Google Faculty Award.
In addition, this material is based on research sponsored by the Air Force Research Laboratory and the Defense Advanced Research Projects Agency under agreement number FA8750-16-2-0204. 
The U.S. Government is authorized to reproduce and distribute reprints for Governmental purposes notwithstanding any copyright notation thereon. 
The views and conclusions contained herein are those of the authors and should not be interpreted as necessarily representing the official policies or endorsements, either expressed or implied, of the Air Force Research Laboratory and the Defense Advanced Research Projects Agency or the U.S. Government.

\bibliographystyle{IEEEtran}
\bibliography{references}

\end{document}